\pdfoutput=1
\documentclass[11pt]{article}

\usepackage[]{acl}

\usepackage{times}
\usepackage{latexsym}

\usepackage[T1]{fontenc}
\usepackage[utf8]{inputenc}

\usepackage{microtype}

\usepackage{makecell}
\usepackage{todonotes}
\usepackage{xcolor}
\usepackage{tikz, pgfplots}
\usetikzlibrary{shapes}
\usepackage{graphicx}
\usepackage{booktabs}
\usepackage{amssymb}
\usepackage{amsmath}
\usepackage{multirow}
\usepackage{float}

\title{ConceptNet infused DialoGPT for Underlying Commonsense Understanding and Reasoning in Dialogue Response Generation}

 \author{Ye Liu$^1$, Wolfgang Maier$^1$, Wolfgang Minker$^2$ \and Stefan Ultes$^1$ \\
 $^{1}$Mercedes-Benz AG, Sindelfingen, Germany \\
 \texttt{\{ye.y.liu,wolfgang.mw.maier,stefan.ultes\}@mercedes-benz.com}\\
 $^{2}$Ulm University, Ulm, Germany \\
 \texttt{ wolfgang.minker@uni-ulm.de}}

\begin{document}
\maketitle
\begin{abstract}

The pre-trained conversational models still fail to capture the implicit commonsense (CS) knowledge
hidden in the dialogue interaction, even though they were pre-trained with an enormous dataset. In order to build a dialogue agent with CS capability, we firstly inject external knowledge into a pre-trained conversational model to establish basic commonsense through efficient Adapter tuning (Section \ref{sec: Commonsense Adapter in DialoGPT}). Secondly, we propose the ``two-way learning'' method to enable the bidirectional relationship between CS knowledge and sentence pairs so that the model can generate a sentence given the CS triplets, also generate the underlying CS knowledge given a sentence (Section \ref{sec: Two-Way Training with CommonGen}). Finally, we leverage this integrated CS capability to improve open-domain dialogue response generation so that the dialogue agent is capable of understanding the CS knowledge hidden in dialogue history on top of inferring related other knowledge to further guide response generation (Section \ref{sec: Commonsense guided DialoGPT Response Model}). The experiment results demonstrate that CS\_Adapter fusion helps DialoGPT to be able to generate series of CS knowledge. And the DialoGPT+CS\_Adapter response model adapted from CommonGen training can generate underlying CS triplets that fits better to dialogue context.

%To be more specific, we firstly infuse DialoGPT with ConceptNet through Adapter layers integrated in the frozen DialoGPT. Secondly, we propose ``two-way learning'' and utilize the CommonGen dataset for the entire DialoGPT+CS\_Adapter model to learn the bidirectional relationship between commonsense knowledge and sentences. We finally enable the commonsense reasoning behind the dialogue through the Commonsense-Dialogues dataset and further train the DialoGPT+CS\_Adapter model to generate a response along with the underlying commonsense triplets. 

\end{abstract}

\section{Introduction}
\label{sec: introduction}

Many pre-trained transformer-based \citep{vaswani2017attention} language models (LMs) have been widely applied in natural language generation (NLG) of spoken dialogue systems (SDS) and shown promising performance. However, the probing experiments about the commonsense (CS) explanation behind the dialogue response generation in \citet{zhou2021probing} demonstrated that pre-trained LMs \citep{radford2019language, zhang2020dialogpt, roller2021recipes, lewis2020bart} fail to capture CS knowledge hidden in dialogue utterances, even though they were already pre-trained with numerous datasets. By contrast, humans generally rely on previous experience and commonsense knowledge to produce coherent responses during the conversation. Hence, improving the CS understanding and reasoning ability of a pre-trained conversational model plays a significant role for the development of SDS.

%However, the pre-traind LMs are still facing the challenging that can not better understand the underlying commonsense (CS) reasoning. 

%The probing experiments about the commonsense explanation in dialogue response generation in \citet{zhou2021probing} demonstrated that pre-trained LMs, which include GPT-2 \citep{radford2019language} based models (DialoGPT \citep{zhang2020dialogpt} and GPT-2) and Seq2Seq based transformer \citet{vaswani2017attention} (BlenderBot \citep{roller2021recipes} and BART \citep{lewis2020bart}), fail to capture the logical relations between commonsense explanations and responses. Even fine-tuning on in-domain data and increasing model sizes did not lead to an understanding of commonsense reasoning for response generation. 

%Meanwhile, the hidden meaning demonstrate not only commonsense capability, but also can further guide much commonsensical response. 
%searches the relative commonsense knowledge to the user utterance before giving a commonsensical response.
% ideal system should not only produce reasonable responses, but also explicitly understand the underlying commonsense reasoning behind the conversation.

\begin{figure}
\begin{tikzpicture}[scale=1]
\footnotesize

\node [] at (0,0) {\includegraphics[width=.03\textwidth]{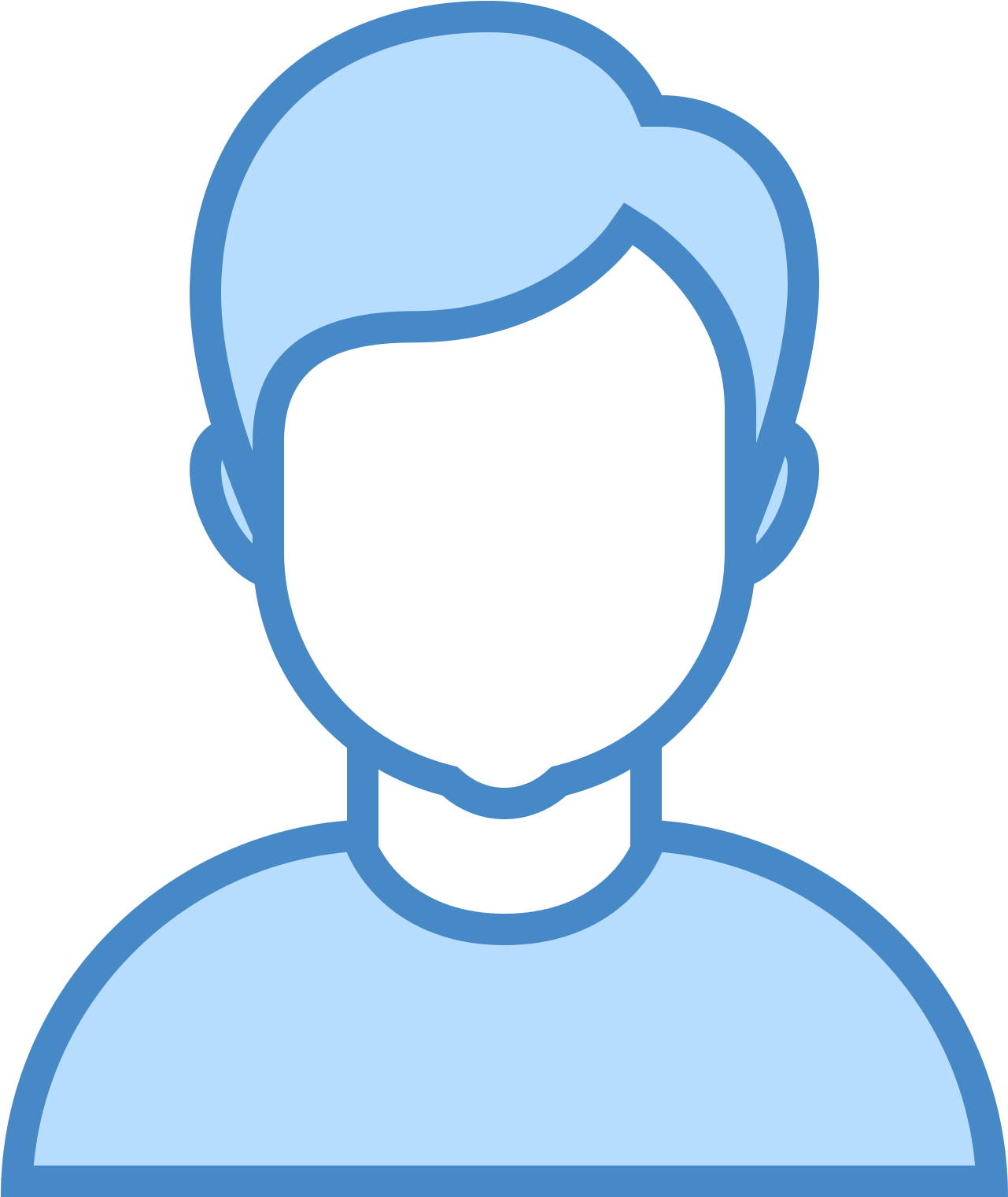}};
\draw [rounded corners] (0.5, -0.3) rectangle (4.5, 0.3);
\node [] at (2.5,0) {I'm on a \textcolor{orange}{diet} to \textcolor{orange}{lose weight}.};

\draw [draw=green, double=green, double distance=2pt, ->] (1.5, -0.4) -- (1.5, -0.9);

\node [] at (4.5, -0.8) {\textbf{Underlying Commonsense Knowledge}};

\draw [rounded corners, dashed] (0.0, -2.2) rectangle (6.5, -1);
\node [] at (7.0, -1.7) {\includegraphics[width=.04\textwidth]{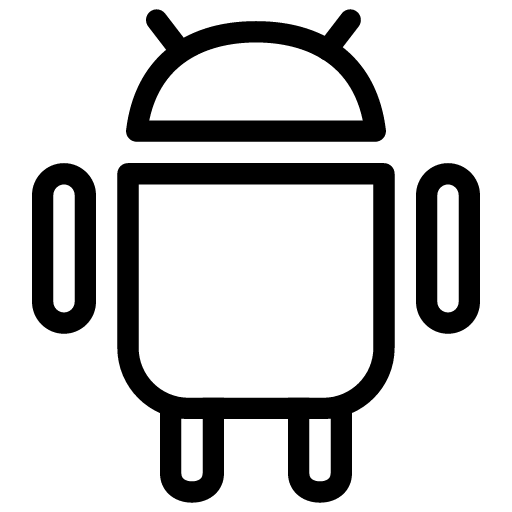}};
\node [] at (3.2,-1.3) {\textcolor{orange}{diet} [has subevent of] \textcolor{orange}{lose weight}};
\node [] at (3.2,-1.6) {\textcolor{orange}{diet} [related to] \textcolor{cyan}{eat}};
%\node [] at (3.2,-1.9) {\textcolor{orange}{diet} [used in context of] \textcolor{cyan}{nutrition}};
%\node [] at (3.2,-2.2) {\textcolor{orange}{diet} [related to] \textcolor{purple}{food}, \textcolor{purple}{food} [has property] \textcolor{cyan}{healthy}};
\node [] at (3.2,-1.9) {\textcolor{orange}{diet} [related to] \textcolor{purple}{food}, \textcolor{purple}{food} [has property] \textcolor{cyan}{healthy}};

\draw [draw=green, double=green, double distance=2pt, ->] (4.5, -2.3) -- (4.5, -2.8);

\node [] at (7, -3.2) {\includegraphics[width=.04\textwidth]{pictures/robot.png}};
\draw [rounded corners] (2.0, -3.5) rectangle (6.5, -2.9);
\node [] at (4.3, -3.2) {Don’t forget to \textcolor{cyan}{eat} more \textcolor{cyan}{healthy}.};
%\node [] at (4.3, -3.6) {and take adequate \textcolor{cyan}{nutrition}.};

\end{tikzpicture}
\caption{\label{fig: The commonsense reasoning between the human-machine interaction.} The human-machine interaction along with the underlying commonsense knowledge/triplets. The key words/concepts are highlighted in \textcolor{orange}{orange} for user utterance and \textcolor{cyan}{blue} for system response, respectively. The words highlighted in \textcolor{purple}{purple} are middle concepts extracted in two-hop searching (Section \ref{subsec: Commonsense Extraction in Concepts}). The dialogue system should have commonsense capability and can generate the underlying commonsense reasoning (in the dotted box) as well as a response.} 
%which includes the commonsense triplets hidden in dialogue context (first one), also commonsense triplets between dialogue context and system response (others),
\end{figure}

%With the help of pre-trained LMs and more available dialogue datasets, SDS can generate replies which are close to satisfying. However, the pre-traind LMs also face a challenge, namely, getting to know the commonsense meaning behind the interactive utterances. \citet{zhou2021probing} probed the commonsense explanation in dialogue response generation and demonstrated that pre-trained models fail to capture the logical relations between commonsense explanations and responses. Even fine-tuning on in-domain data and increasing model sizes did not lead to an understanding of commonsense reasoning for response generation. 

The ideal dialogue agent should be capable of capturing the underlying commonsense knowledge besides generating a reasonable response, as shown in Figure~\ref{fig: The commonsense reasoning between the human-machine interaction.}. In order to achieve that and enable the commonsense capability of a pre-trained conversation model, we have the following contributions in this work:
\begin{enumerate}
    \item 
    We firstly induce commonsense knowledge into a pre-trained conversational model so that it is ultimately able to generate series of commonsense triplets (please refer to Table \ref{tab: use cases for the cs adapter integration}). This is done by integrating Adapter \citep{houlsby2019parameter, pfeiffer2021adapterfusion} layers into the pre-trained DialoGPT \citep{zhang2020dialogpt} and leveraging the commonsense knowledge in ConceptNet \citep{liu2004conceptnet} to train the Adapter layers infused in the frozen DialoGPT. 
    %Thus, the DialoGPT after CS\_Adapter integration is capable of generating series of commonsense knowledge shown in Table \ref{tab: use cases for the cs adapter integration}.
    
    \item
    Secondly, the model is expected to capture the bidirectional relationship between commonsense knowledge and sentence pairs, so that it can both generate sentences given commonsense triplets and underlying commonsense triplets given a sentence (please refer to Table \ref{tab: use cases for the commongen bridge model}). Hence, we extract the CS knowledge between concepts in CommonGen dataset \citep{lin-etal-2020-commongen} and continually train the entire model: DialoGPT+CS\_Adapter, through the proposed `two-way learning''. 
    
    \item
    To finally enable the commonsense ability in dialogue response model that can understand underlying commonsense knowledge and infer related other knowledge for the response generation
    (please refer to Table \ref{tab: use cases for the dialogpt+cs adapter generation mode}), we further utilize the Commonsense-Dialogues \citep{zhou-etal-2021-commonsense} dataset to continually train the DialoGPT+CS\_Adapter model.
    
\end{enumerate}

\section{Related Works}
\label{sec: related works}

Pre-trained LMs have been frequently used in task-oriented and chit-chat response generation in SDS, also gained impressive performance.
\citet{peng2020few, chen2020few} and \citet{peng2021soloist} introduced few shot and end-to-end task-oriented NLG with pre-trained GPT-2 \citep{radford2019language}. \citet{kale2020template} utilized the pre-trained T5 encoder-decoder model \citep{raffel2020exploring} to re-write the template guided task-oriented text generation. \citet{zandie2020emptransfo} and \citet{lin2020caire} utilized pre-trained GPT \citep{radfordimproving} for creating an empathetic chatbot. In our work, we not only utilize pre-trained LM, but also improve the commonsense reasoning ability of DialoGPT.

Even though the large-scale pre-trained LMs have demonstrated impressive performance on multiple generation tasks, building a generation model with commonsense to compose realistically plausible sentences remains challenging. With the new benchmark dataset CommonGen released in \citet{lin-etal-2020-commongen}, many works haven been presented in these two years for testing the ability of generative commonsense reasoning. \citet{fan2020enhanced} enhanced the retrieved prototype into BART \citep{lewis2020bart} and combined scaling module and prototype position indicator to better utilize the scenario knowledge of prototype for CommonGen text generation. \citet{liu2021kg} proposed KG-BART, where the common sense knowledge graph was incorporated into the pre-trained BART \citep{lewis2020bart} both encoder and decoder side to promote the ability of commonsense reasoning for text generation. \citet{wang2021retrieval} firstly retrieved prototype sentence candidates by a concept matching retriever and a trainable sentence retriever, then used them as auxiliary input to further boost the common sense generation by adopting pre-trained T5 \citep{raffel2020exploring} encoder-decoder model. \citet{feng2021retrieve} leveraged the information contained in images to enhance the common sense text generation. To be more precise, they retrieved images for each concept set and generated a caption for the image via a pre-trained image captioning model, then the captions as augmented inputs were utilized to boost common sense generation. However, these works only focus on the common sense sentence in daily scenario, the work about common sense guided responses generation in spoken dialogue system is still underexplored. Our work tries to leverage the common senses reasoning capability in the dialogue generation systems.
% where given a set of common concepts (e.g., {dog, frisbee, catch, throw}), the task is to generate a coherent sentence describing an everyday scenario using these concepts (e.g., ``a man throws a frisbee and his dog catches it''); 

In recent years, many works also tried to infuse external knowledge into pre-trained models to improve the performance of downstream tasks. \citet{lauscher2020common} investigated the Adapter-based knowledge injection into pre-trained BERT \citep{kenton2019bert} and improved the language understanding ability. \citet{liu2021empathetic} integrated commonsense knowledge and emotional concepts into a pre-trained encoder-decoder architecture to improve emotion recognition ability and produce more empathetic responses. Our work mainly enables the commonsense reasoning capability of pre-trained DialoGPT \citep{zhang2020dialogpt} through light Adapter integration and ConceptNet.

\section{ConceptNet, CommonGen and Commonsense-Dialogues}
\label{sec: commonsense dialogues and conceptnet}
The commonsense knowledge graph ConceptNet \citep{liu2004conceptnet}, the CommonGen \citep{lin-etal-2020-commongen} and the dialogue dataset Commonsense-Dialogues \citep{zhou-etal-2021-commonsense} are used in this work and are briefly introduced in this section.

\paragraph[]{ConceptNet\footnote{https://conceptnet.io/}} is a large-scale and multilingual commonsense knowledge graph that describes general human knowledge in natural language. It comprises $5.9$M assertions, $3.1$M concepts and $38$ relations. The nodes in ConceptNet are concepts and the edges are relations. Each (head concept, relation, tail concept) triplet is an assertion. Each assertion is associated with a weight. The Table \ref{tab: example assertion in the ConceptNet} shows several assertions along with corresponding relation in the ConceptNet. In our work, the ConceptNet is firstly used to infuse the commonsense knowledge into the pre-trained DialoGPT through Adapter tuning (Section \ref{sec: Commonsense Adapter in DialoGPT}) and further used to extract the commonsense triplets between key concepts (Section \ref{subsec: Commonsense Extraction in Concepts} and \ref{subsec: keywords and commonsense extraction}) for the entire DialoGPT+CS\_Adapter training.

\begin{table}
\footnotesize
\begin{center}
\begin{tabular}{cccc}
    \toprule
    head concept & relation  & tail concept  & weight \\
     
   \midrule
   loneliness  & CausesDesire  & socialize  & $3.464$  \\
   plate  & AtLocation  & restaurant & $2.0$  \\
   program  & CreatedBy  & programmer & $6.633$  \\

   \bottomrule
\end{tabular}
\end{center}
\caption{\label{tab: example assertion in the ConceptNet} The commonsense assertion examples with different relations in ConceptNet. The weight represents the strength with which the edge expresses this assertion and usually in the $[1, 10]$ interval.}
\end{table}

\paragraph[]{CommonGen\footnote{https://inklab.usc.edu/CommonGen/}} is a challenging dataset, which includes concepts-sentences pairs shown in Table \ref{tab: example in CommonGen along with extracted CS triplets}. This dataset is widely used to explicitly test machines for the ability of generative commonsense reasoning. The sentences in CommonGen generally describe the everyday scenarios using these concepts. In this work, CommonGen is utilized  to train the DialoGPT+CS\_Adapter through `two-way learning'' (Section \ref{subsec: Two-Way training}) with the extracted commonsense triplets and sentence pair (Section \ref{subsec: Commonsense Extraction in Concepts}) to learn the bidirectional relationship between commonsense knowledge and sentence pairs.

\paragraph[]{Commonsense-Dialogues\footnote{https://github.com/alexa/commonsense-dialogues}} was released in \citet{zhou-etal-2021-commonsense} and is a crowdsourced corpus of around $11$K dialogues based on SocialIQA \citep{sap2019socialiqa} event prompt. Commonsense-Dialogues involve much commonsense reasoning in the dialogue and help models produce more commonsense responses. The Commonsense-Dialogues is utilized finally to activate the commonsense reasoning behind the dialogue for the response generation model (Section \ref{sec: Commonsense guided DialoGPT Response Model}).

%The Table \ref{tab: statistic information of Commonsense-Dialogues} shows some statistics about Commonsense-Dialogues dataset.

%\begin{table}
%\footnotesize
%\begin{center}
%\begin{tabular}{l|ccc}
%    \toprule
%    state & train  & valid  & test \\
%     
%   \midrule
%   \# of dialogues  & $9058$  & $1157$  & $1158$  \\
%   average \# of turns in a dialogues  & $5.72$  & $5.72$  & $5.71$  \\
%   average \# of words in a turn  & $12.4$  & $12.4$  & $12.2$  \\
%   
%   \bottomrule
%\end{tabular}
%\end{center}
%\caption{\label{tab: statistic information of Commonsense-Dialogues} Some statistics about Commonsense-Dialogues dataset.}
%\end{table}

\section{Adapter tuning in DialoGPT with ConceptNet}
\label{sec: Commonsense Adapter in DialoGPT}

The empirical study in \citet{zhou-etal-2021-commonsense} demonstrated that commonsense knowledge helps to boost response generation. To enable the commonsense capability, we infuse the commonsense knowledge into the pre-trained conversational model DialoGPT \citep{zhang2020dialogpt}, through efficient Adapter tuning. That means, only the parameters of the Adapter layers are fine-tuned, while the parameters of the pre-trained DialoGPT remain frozen. As a result, it is efficient and lightweight for commonsense knowledge infusion. In this work, we utilize the AdapterHub \citep{pfeiffer2020adapterhub} to integrate the Adapter layers into DialoGPT and then train the Adapter layers with the synthetic commonsense corpus from ConceptNet.

%In this section, we will introduce commonsense corpus synthetically collected from ConceptNet and details of Adapter training with the synthetic corpus.

\subsection{Corpus collection in ConceptNet}
\label{subsec: Biased Random Walks in ConceptNet}
We adapt the work from \citet{perozzi2014deepwalk, lauscher2020common} and induce a synthetic corpus from ConceptNet through bias random walking \cite{grover2016node2vec} its graph. 
 
%In order to simplify the ConceptNet traversing process, we only retain the commonsense assertion, which head concept or tail concept exists in Commonsense-Dialogues vocabulary. 

%We also found that using the Glove cosine similarity as weight is beneficial to collecting much reasonable corpus.

Given a source concept, we simulate a random walk of fixed length $l$. Let $c_{i}$ represents the $i$th concept in the walk. Starting with $c_{0}$, we firstly sample one concept as $c_{1}$ from its neighbors based on the normalized transition probability, which is shown in
Equation \ref{equ: normalized transition probability}. The $\pi_{vx}$ denotes unnormalized transition probability. $Z$ means the size of neighbors and $G$ represents the entire ConceptNet Graph.
\begin{equation}
    P(c_{i} = x | c_{i-1} = v) = \frac{\pi_{vx}}{Z}, \hspace{0.3cm} if \hspace{0.1cm} (v, x) \in G \;
\label{equ: normalized transition probability}
\end{equation}

One thing matters here, even though ConceptNet provides weight for the assertions, they do not seem to be assigned with high confidence\footnote{https://github.com/commonsense/conceptnet5/issues/152}. Hence, we use the cosine similarity between head concept and tail concept embedded by Glove vectors \citep{pennington2014glove} to replace the original weight\footnote{In this work, we remove the assertions with Glove cosine similarity between nodes $v$ and $x$ less than $0$, also the assertions with original weight less than $1$.}. Hence, the $w_{vx}$ in Equation \ref{equ: unnormalized transition probability} denotes the Glove cosine similarity between head concept $v$ and tail concept $x$.

\begin{equation}
\begin{split}
    \pi_{vx} & = w_{vx}, \hspace{2cm} if \hspace{0.1cm} i=1 \\
            & = \alpha_{pq(t,x)} * w_{vx}, \hspace{0.5cm} otherwise
\end{split}
\label{equ: unnormalized transition probability}
\end{equation}

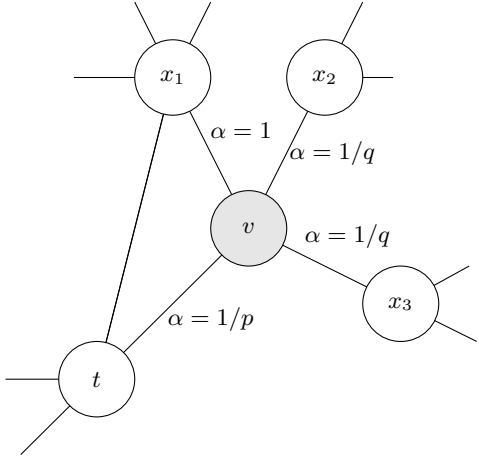
\begin{figure}
\begin{tikzpicture}[scale=1]
\footnotesize

\draw[color=black](0,0) -- (2,-1);
\node [] at (1.3,-0.1) {$\alpha=1/q$};
\draw[color=black](0,0) -- (1,2);
\node [] at (1.1,1) {$\alpha=1/q$};
\draw[color=black](0,0) -- (-1,2);
\node [] at (-0.1,1.3) {$\alpha=1$};
\draw[color=black](0,0) -- (-2,-2);
\node [] at (-0.5,-1.2) {$\alpha=1/p$};
\draw[color=black](-1,2) -- (-2,-2);

\draw[color=black](1,2) -- (1.5,3);\draw[color=black](1,2) -- (1.9,2);
\draw[color=black](2,-1) -- (3,-1.5);\draw[color=black](2,-1) -- (2.9,-0.5);
\draw[color=black](-1,2) -- (-2,-2);
\draw[color=black](-1,2) -- (-1.5,3);\draw[color=black](-1,2) -- (-0.5,3);\draw[color=black](-1,2) -- (-2.3,2);
\draw[color=black](-2,-2) -- (-3,-3);
\draw[color=black](-2,-2) -- (-3.2,-2);

\filldraw[color=black, fill=gray!20](0,0) circle (0.5);
\node [] at (0,0) {$v$};    
\filldraw[color=black, fill=white](2,-1) circle (0.5);
\node [] at (2,-1) {$x_3$};
\filldraw[color=black, fill=white](1,2) circle (0.5);
\node [] at (1,2) {$x_2$};
\filldraw[color=black, fill=white](-1,2) circle (0.5);
\node [] at (-1,2) {$x_1$};
\filldraw[color=black, fill=white](-2,-2) circle (0.5);
\node [] at (-2,-2) {$t$};

\end{tikzpicture}
\caption{\label{fig: Illustration of the random walk procedure in ConceptNet.} Illustration of the biased random walks (proposed in \citet{perozzi2014deepwalk}) through ConceptNet. The walk just transitioned from concept $t$ to concept $v$ and is now evaluating its next concept $v$. Edge labels indicate search biases $\alpha$.} 
\end{figure}

Since starting searching $c_{2}$, we will bias the random walks by introducing search bias $\alpha$. The work in \citet{perozzi2014deepwalk} defines two parameters $p$ and $q$, which is shown in Figure \ref{fig: Illustration of the random walk procedure in ConceptNet.}, to control how fast the walk explores and how far away the next concept is from the last concept. Consider the walk that just traversed edge $(t, v)$ and now resides at concept $v$ (Figure \ref{fig: Illustration of the random walk procedure in ConceptNet.}). Now, we set the unnormalized transition probability to $\pi_{vx}=\alpha_{pq(t,x)} * w_{vx}$ (see Equation \ref{equ: unnormalized transition probability}), where
\begin{equation}
\begin{split}
    \alpha_{pq}(t, x) & = \frac{1}{p} \hspace{0.5cm} if \hspace{0.1cm} d_{tx}=0 \\
                      & = 1  \hspace{0.5cm}  if \hspace{0.1cm} d_{tx}=1 \\
                      & = \frac{1}{q} \hspace{0.5cm} if \hspace{0.1cm} d_{tx}=1  \\
\end{split}
\label{equ: search bias}
\end{equation}
and $d_{tx}$ denotes the shortest path distance between $t$ and $x$ and $d_{tx}$ must be one of ${0, 1, 2}$. By setting $q > 1$, the random walk is biased towards nodes close to concept $t$. Setting $p > \max(q, 1)$ can ensure that we are less likely to sample an already visited concept. We define $p=2.0, q=1.5$ in this work to enable the walking through much relative concepts to the previous concept, meanwhile avoid repeating it. The walking will be terminated, until to length $l$ or there is no path to traverse. We set $l=10$ and traverse the ConceptNet graph twice to collect the corpus. Finally, we collected $359,421$ data points and split them to train/valid/test as 80\%/10\%/10\%. After simple dataset processing, the collected corpus is like the data in Table \ref{tab: example for collected data and inputt prompts}.

%``autobraking [related to] automatic, automatic [derived from] auto, auto [related to] automobile, automobile [related to] car, car [related to] cable''

\begin{table*}
\footnotesize
\begin{center}
\begin{tabular}{cc}
    \toprule
   data & autobraking [related to] automatic, automatic [derived from] auto, auto [related to] automobile, automobile [related to] car  \\ \hline
   \makecell[l]{prompt \\ templates} & \makecell[l]{<|commonsense|> autobraking [related to] \\ 
                <|commonsense|> autobraking [related to] automatic, \\
                <|commonsense|> autobraking [related to] automatic, automatic [derived from]\\
                <|commonsense|> autobraking [related to] automatic, automatic [derived from] auto,} \\
   \bottomrule
\end{tabular}
\end{center}
\caption{\label{tab: example for collected data and inputt prompts} One example in collected corpus from ConceptNet (Section \ref{subsec: Biased Random Walks in ConceptNet}) and four prompt templates are randomly chosen as input during CS\_Adapter training (Section \ref{subsec: Commonsense Adapter Training}).}
\end{table*}

\subsection{Commonsense Adapter Training}
\label{subsec: Commonsense Adapter Training}
To enable the commensense ability, the collected corpus in Section \ref{subsec: Biased Random Walks in ConceptNet} is utilized to train the commonsense Adapter\footnote{In the early stage of this work, we tried Houlsby Adapter \citet{houlsby2019parameter} and Pfeiffer Adapter \citep{pfeiffer2021adapterfusion} both. Houlsby has slightly better performance, hence, we choose the widely used Houlsby Adapter in the following work.} (CS\_Adapter) layers infused in the frozen DialoGPT \citep{zhang2020dialogpt}.
In order to ensure the success of the CS\_Adapter training, we propose the following tips:
\begin{itemize}
    \item 
    We add one special token ``<|commonsense|>'' to the DialoGPT tokenizer and insert this special token at the beginning of every input prompt (See Table \ref{tab: example for collected data and inputt prompts}). Through ``<|commonsense|>'', the model can distinguish the commonsense triplets from the normal text.
    %(Section \ref{sec: Two-Way Training with CommonGen}) and dialogue utterances (Section \ref{sec: Commonsense guided DialoGPT Response Model}).
    
    \item
    We create a mapping table between the relations in ConceptNet and natural language (NL) phrases, which is shown in Table \ref{tab: mapping between assertion relation and natural language phrase}. Furthermore, we utilize the special character $[ ]$ along with the relation phrases. Our experiments show that $[ ]$ highly helps the model to distinguish the relation from normal words/concepts.
    
    \item
    Given the auto-regressive property of DialoGPT, four prompt templates shown in Table \ref{tab: example for collected data and inputt prompts} are proposed and randomly chosen as input to guide the generation of commonsense knowledge. Meanwhile, the prompt inputs of training date is re-sampled every $3$ epochs.
    
\end{itemize}
We train CS\_Adapter layers with batch size: $64$, learning rate: $5\mathrm{e}{-5}$ and best model is saved at epoch $20$. During decoding, we mix top-K sampling of $5$ and top-p (nucleus) sampling of $0.9$ \citep{holtzman2019curious}, and we generate $5$ examples for every test prompt input (Same decoding strategy in Section \ref{subsec: Two-Way training} and \ref{subsec: DialoGPT+CS Adapter Training}).

\begin{table}
\centering
\scalebox{0.95}
{\begin{tabular}{ccc}
    \toprule
    &  annotator 1 & annotator 2 \\
    
    \hline
    yes \emph{vs}  no &  87 \emph{vs}  13 & 88 \emph{vs}  12\\

    positive agreement & \multicolumn{2}{c}{93.71\%} \\

    %negative agreement & \multicolumn{2}{c}{56\%} \\

    %cohen's kappa & \multicolumn{2}{c}{49.73\%} \\
    \bottomrule
    
\end{tabular}}
\caption{The human assessment results on generated commonsense triplets that do not officially exist in ConceptNet. The high positive agreement demonstrates that even though the generated CS triplets do not officially exist in ConceptNet, there is still high potential that they make sense for humans.}
\label{tab: The human assessment results on commonsense triplets generated from DialoGPT+CS Adapter.}
\end{table}

\section{Two-Way Learning with CommonGen}
\label{sec: Two-Way Training with CommonGen}

After CS\_Adapter infusion, the DialoGPT model can generate series of commonsense triplets (Table \ref{tab: use cases for the cs adapter integration}). Now, we except the model to capture the ``bidirectional'' relationship between commonsense knowledge and sentence pairs. That means, it can generate a sentence given some commonsense triplets; meanwhile, is also able to generate the underlying commonsense knowledge given a sentence. This is what the ``bidirectional'' means here. As a result, this step is as a bridge to the dialogue response model that can generate underlying commonsense knowledge given dialogue context and a response given the generated CS knowledge. Hence, we extract the commonsense triplets between the concepts in CommonGen \citep{lin-etal-2020-commongen} and leverage the CommonGen for the bridge dataset to further train the DialoGPT+CS\_Adapter with our proposed ``two-way learning''.

\subsection{Commonsense Extraction in Concepts}
\label{subsec: Commonsense Extraction in Concepts}
In the CommonGen, the key concepts (also keywords) are already provided for every sentence. Hence, we need to further extract the underlying relational commonsense knowledge between the concepts from ConceptNet. Not like \citet{zhou-etal-2021-commonsense}, only one-hop triplets were filtered; however, one-hop and two-hop triplets are both searched in this work. For the commonsense triplets filtering in ConceptNet, we have the following strategies:
\begin{enumerate}
    \item 
    We traverse every two concepts in the concept set, and firstly search if there is an one-hop triplet in ConceptNet for the two concepts; if not, then activate the two-hop searching. 
    
    Like the first example in Table \ref{tab: example in CommonGen along with extracted CS triplets}, there is the one-hop triplet ``surfer [related to] surf'' for the concept ``surfer'' and ``surf''. Even though there is no one-hop triplet for ``ocean'' and ``surf'' in ConceptNet, we find the two-hop triplet ``surfing [has prerequisite] ocean, surfing [related to] surf'' with the middle concept ``surfing'' through two-hop searching.
    
    \item
    When activating two-hop searching, there are generally many two-hop triplets for any two concepts. Hence, we propose three thresholds\footnote{The three settings are empirically determined and highly improved the extraction quality.} to pick most relevant commonsense triplet out. 
    Firstly, we need to make sure the cosine similarity between the two concepts embedded by Glove vectors \citep{pennington2014glove} is not less than $0.3$.
    Secondly, we need to make sure the cosine similarity between the middle concept and at least one of the two concepts is larger than $0.5$.
    Finally, we only select the commonsense triples with the highest weight score.

\end{enumerate}
The Table \ref{tab: example in CommonGen along with extracted CS triplets} shows two CommonGen dataset examples along with the extracted commonsense knowledge. And the selected two-hop commonsense triplets are highlighted in \textcolor{red}{red}.

\begin{table*}
\centering
\scalebox{0.9}
{\begin{tabular}{cc}
    \toprule
    concepts & ocean | surfer | surf \\
    
    %\hline
   extracted CS triplets  & \textcolor{red}{surfing [has prerequisite] ocean, surfing [related to] surf;} surfer [related to] surf \\
     
   %\hline
   sentences  & The ocean is where surfers go to surf. // A surfer surfing in the ocean. \\
                
    \midrule
    concepts & table | burger | eat \\
   
   %\hline
   extracted CS triplets & table [related to] eat; \textcolor{red}{burger [is a] food, food [makes someone want] eat;} \\
   %\hline
   sentences  & They eat burgers at the dinner table. // The man sat at the table to eat the burger. \\

   \bottomrule
\end{tabular}}
\caption{\label{tab: example in CommonGen along with extracted CS triplets} Two data examples in CommonGen along with the extracted commonsense (CS) triplets from ConceptNet. The two-hop triplets are highlighted in \textcolor{red}{red}. Every example is attached with two sentences here and separated with //.}
\end{table*}

\subsection{Two-Way Learning}
\label{subsec: Two-Way training}

The Table \ref{tab: statistic information of CommonGen with extracted commonsense knowledge} shows the statistics of extracted CommonGen dataset with commonsense triplets, which are utilized to continually train the entire DialoGPT+CS\_Adapter for learning the bidirectional relationship between CS knowledge and sentences.

In this step, we propose the ``two-way learning'' to train the DialoGPT+CS\_Adapter model. That means, for every CS\_triplets-sentence pair, which is shown in Table \ref{tab: example in CommonGen along with extracted CS triplets}, when we input the extracted CS triples to the DialoGPT+CS\_Adapter model, the output is the sentence; on the contrary, when the input is the sentence, the model will output the extracted CS triples. Through this 'two-way learning', the model can learn to generate a sentences given the CS knowledge and the underlying commonsense triplets given a sentence (Table \ref{tab: use cases for the commongen bridge model}). During the two-way training, all the parameters in DialoGPT+CS\_Adapter model are activated and updated with batch size: $16$, learning rate: $5\mathrm{e}{-5}$, and best model is saved at epoch $1$ with early stopping.
%This training is used as a bridge to the commonsense guided response generation model that is capable of generating the underlying commonsense knowledge behind the dialogue interaction, meanwhile a commonsense response given the former generated commonsense reasoning.

\begin{table}
\footnotesize
\begin{center}
\begin{tabular}{l|ccc}
    \toprule
    state & train  & valid  & test \\
     
   \midrule
   \# of triplet-sentence pair  & $10,880$  & $1645$  & $677$  \\
   average \# of CS triplets  & $4.61$  & $4.89$  & $4.70$  \\

  \bottomrule
\end{tabular}
\end{center}
\caption{\label{tab: statistic information of CommonGen with extracted commonsense knowledge} The statistics about CommonGen dataset with extracted commonsense triplets.}
\end{table}

\section{Commonsense guided DialoGPT Response Model}
\label{sec: Commonsense guided DialoGPT Response Model}
After CS\_Adapter integration, the DialoGPT can generate series of commonsense triplets. Then the DialoGPT+CS\_Adapter is further trained to learn the bidirectional relationship between commonsense knowledge and daily sentence pairs through ``two-way learning'' with CommonGen dataset. Now, we expect the model to be a dialogue response generation model with commonsense capability for the final goal of this work. To be more specific, it can generate the commonsense knowledge hidden in the dialogue interaction, meanwhile a reasonable response given the generated CS triplets and dialogue context. Hence, adapted the DialoGPT+CS\_Adapter model after ``two-way learning'', we leverage a commonsense focused dialogue dataset: Commonsense-Dialogues \citep{zhou-etal-2021-commonsense} and further train the DialoGPT+CS\_Adapter model to generate the commonsense reasoning along with a response given the dialogue context.

\subsection{Keywords and Commonsense Extraction}
\label{subsec: keywords and commonsense extraction}
Firstly, we need to extract the key words/concepts from the Commonsense-Dialogues sentences, because it is not like CommonGen dataset where the concepts in the sentences are already provided. For the keywords extraction, we adapt the work from \citet{tang2019target} and \citet{zhong2021keyword}, use TF-IDF and Part-Of-Speech (POS) features to select the keywords. Afterwards, the same commonsense extraction method (Section \ref{subsec: Commonsense Extraction in Concepts}) is applied to pick out the commonsense triplets between the keywords. During the commonsense extraction, like the Figure \ref{fig: The commonsense reasoning between the human-machine interaction.} shows, not only the commonsense triplets between the keywords in dialogue contexts, but also between the keywords in dialogue context and system response, both are extracted for every dialogue. The Table \ref{tab: statistic about the extratced CS triplets in Commonsense-Dialogues dataset} shows the statistic of the extracted commonsense triplets hidden in Commonsense-Dialogue dataset. %Because we expect the model that can not only capture the commonsense reasoning hidden in dialogue context, but also generate the commonsense knowledge that exerts keywords that can further guide the response generation.

%In addition, when there is no direct relation (also one-hop neighborhood) between the keywords in utterances, we will try to search the two-hop reasoning. Such as: there is no relation between keyword ``challenge'' and ``achieve''. By two-hop searching, we could extract ``challenge [typically located at] work; work [related to] achieve'' to enrich the commonsense knowledge in the utterances.

\begin{table*}
\centering
\begin{tabular}{c|ccc}
\toprule
    extracted CS triplets &  train & valid & test \\
\midrule
    only hidden in dialogue contexts (\%) & 47.05  & 46.24   &  46.08   \\
    only hidden in dialogue context and response (\%) &  24.14  &  25.68  & 25.41  \\
    hidden in above-mentioned both sides (\%) &  28.81  & 28.07  &  28.52 \\
\bottomrule  
\end{tabular}
\caption{The statistic of the extracted commonsense triplets in Commonsense-Dialogues dataset.}
\label{tab: statistic about the extratced CS triplets in Commonsense-Dialogues dataset}
\end{table*}

\subsection{DialoGPT+CS\_Adapter Response Training}
\label{subsec: DialoGPT+CS Adapter Training}
In this step, we adapt the model from ``two-way learning'' with CommonGen and continually train DialoGPT+CS\_Adapter model with Commonsense-Dialogues \citep{zhou-etal-2021-commonsense} to enable the commonsense understanding and reasoning behind dialogue interaction for open-domain response generation. The dialogue context is as input, the extracted commonsense triplets and response as label to guide the training. In this step, we add two new tokens: ``[USER]'' and ``[SYSTEM]'' to distinguish the user utterance from system response. During training, maximal $3$ turns' dialogue context are taken into account for memory-efficient. We train the entire model: DialoGPT+CS\_Adapter with batch size: $16$, learning rate: $5\mathrm{e}{-5}$, and best model is saved at epoch $5$ with early stopping.

\begin{table*}
\footnotesize
\begin{center}
\begin{tabular}{cccc}
    \toprule
    model & perplexity$\downarrow$ & concepts Acc (\%) & assertion Acc (\%)\\
   \midrule
   DialoGPT+CS\_Adapter integration & - & $56.88$ & $47.29$ \\
   \hline
   DialoGPT baseline (w/o) CS\_Adapter integration & $1.405$ & -  & - \\
   DialoGPT+CS\_Adapter (w/o) two-way learning  &  $1.365$  & $62.43$  & $45.27$\\
   DialoGPT+CS\_Adapter (final) &  $1.364$  & $63.66$  & $47.28$  \\
   \bottomrule
\end{tabular}
\end{center}
\caption{\label{tab: performance comparison DialoGPT VS different DialoGPT+CS Adapter} The automatic metrics of DialoGPT after CS\_Adapter integration (Section \ref{sec: Commonsense Adapter in DialoGPT}). And the performance comparison of different response generation models: DialoGPT baseline without CS\_Adapter, DialoGPT+CS\_Adapter without two-way learning (Section \ref{sec: Two-Way Training with CommonGen}) and DialoGPT+CS\_Adapter adapted from two-way learning (Section \ref{sec: Commonsense guided DialoGPT Response Model}).}
\end{table*}

\begin{table*}
\footnotesize
\centering
\begin{tabular}{c|c|c}
    \toprule
    models comparison & CS triplets(\%) & response(\%)  \\ %& CS guidance over response(\%) 
    \midrule
    DialoGPT+CS\_Adapter (final) \emph{vs} (w/o) two-way learning & 23 \emph{vs}  20 & 21 \emph{vs}  28  \\ 
    %& 8 \emph{vs}  11
    
    \hline
    DialoGPT+CS\_Adapter (final) \emph{vs} DialoGPT baseline  &  -  & 20 \emph{vs}  28 \\ %& - 
    \bottomrule
\end{tabular}
\caption{The human A/B evaluation on DialoGPT+CS\_Adapter \emph{vs} \{(w/o) two-way learning, DialoGPT baseline\}.}
\label{tab: human evaluation on different dialogpt+cs_adapter moodels and baselines}
\end{table*}

\section{The Experiment Results}
\label{sec: the experiment results}
This section introduces the experiment results of this work through automatic metrics, human assessment and use cases. We will evaluate the performance of DialoGPT after CS\_Adapter integration firstly, which is the basis of the commonsense guided dialogue response model. Afterwards, we will compare the performance of different response generation models.

\subsection{Evaluation of Commonsense Adapter}
\label{subsec: Evaluation of Commonsense Adapter}
This section will present the evaluation results of CS\_Adapter integration (Section \ref{sec: Commonsense Adapter in DialoGPT}) in DialoGPT.

We firstly propose two automatic metrics to evaluate the performance of CS\_Adapter infused in DialoGPT for generating series of commonsense triplets. One is $\emph{concepts accuracy}$, which represents the proportion of generated (head concept, tail concept) that exists in ConceptNet without considering if the generated relation is officially correct. Because there is not one unique relation for (head concept, tail concept), so the concepts accuracy can identify to some extend that model can understand the head concept and tail concept is related. Another is $\emph{assertion accuracy}$, which represents the proportion of generated (head concept, relation, tail concept) that officially exists in ConceptNet. The automatic metric results of DialoGPT+CS\_Adapter integration in the first row of Table \ref{tab: performance comparison DialoGPT VS different DialoGPT+CS Adapter} show that only about half of  generated commonsense triplets exist officially in ConceptNet with $56.88\%$ concepts accuracy and $47.29\%$ assertion accuracy.

In order to further prove that even if the generated commonsense triplets do not officially exist in ConceptNet, they still make sense for humans, we hire two Master students with computational linguistic background to manually evaluate the generated commonsense assertions which do not officially exist in ConceptNet. We pick out $50$ generated assertions that (head concept, tail concept) that exists in ConcepeNet, while the predicted relation is not officially correct; and $50$ generated assertions that (head concept, tail concept) that do not officially exist in ConcepNet. Two annotators were asked to assess that generated (head concept, relation, tail concept) is reasonable or not by answering ``yes'' or ``no''. The human assessment results shown in Table \ref{tab: The human assessment results on commonsense triplets generated from DialoGPT+CS Adapter.} support our initial assumption that even though the generated commonsense triplets do not officially exist in the ConceptNet, they are potentially high reasonable for humans with up to $93.17\%$ positive agreement.

Additionally, we also show several use cases in Table \ref{tab: use cases for the cs adapter integration} to demonstrate that the DialoGPT after CS\_Adapter fusion have the commonsense ability and is able to generate series of reasonable commonsense triplets. This is the basis for the dialogue response model with commonsense capability.

\subsection{Evaluation of Commonsense guided Response Model}
\label{subsec: Evaluation of Commonsense guided Response Model}
This section will introduce the performance comparison of different dialogue response generation models. In Table \ref{tab: performance comparison DialoGPT VS different DialoGPT+CS Adapter}, the DialoGPT baseline represents the DialoGPT is directly fine-tuned with Commonsense-Dialogues dataset for response generation without Adapter tuning and ``two-way learning''; the DialoGPT+CS\_Adapter without (w/o) two-way learning means that the DialoGPT after CS\_Adapter fusion is further trained with Commonsense-Dialogues dataset; the final DialoGPT+CS\_Adapter response model is adapted from two-way learning with CommonGen dataset and then further trained with Commonsense-Dialogues dataset. Hence, the former DialoGPT baseline model can only generate response; while the latter two DialoGPT+CS\_Adapter models can both generate commonsense triplets hidden in dialogue interaction as well as a response. Besides automatic metrics in Table \ref{tab: performance comparison DialoGPT VS different DialoGPT+CS Adapter}, three annotators with computational linguistic background are hired for human evaluation. In this human evaluation task, we randomly sample $100$ dialogues each for DialoGPT+CS\_Adapter (final) $\emph{vs}$ \{DialoGPT+CS\_Adapter (w/o) two-way learning, DialoGPT baseline\}. For the former comparison, the annotators are asked to choose the better commonsense triplets and better response respectively. In the latter comparison, the annotators are asked to choose the better response. They can either choose one better option or select $\emph{tie}$ when the provided options are either both good or both bad. Finally, the median score of three annotations is computed for the results comparison shown in Table \ref{tab: human evaluation on different dialogpt+cs_adapter moodels and baselines}.

The perplexity \citep{serban2015hierarchical} values are utilized to measure the high-level general quality of the generation model. Lower perplexity of DialoGPT+CS\_Adapter in Table \ref{tab: performance comparison DialoGPT VS different DialoGPT+CS Adapter} indicates its better generalization performance. Compared with DialoGPT+CS\_Adapter (w/o) two-way learning, the final DialoGPT+CS\_Adapter response model has higher concepts accuracy and assertion accuracy, which demonstrates the DialoGPT+CS\_Adapter response model benefits from the two-way learning with CommonGen dataset and can generate better commonsense triplets that fit well to the dialogue context. This is further underpinned by human evaluation results on CS triplets in Table \ref{tab: human evaluation on different dialogpt+cs_adapter moodels and baselines}. However, the results of human evaluation on response show that the generated response of DialoGPT+CS\_Adapter has worse performance than the other two models. The possible reason is that the extracted commonsense knowledge for Commonsense-Dialogues (Section \ref{subsec: keywords and commonsense extraction}) are not as perfect as in the Figure \ref{fig: The commonsense reasoning between the human-machine interaction.} shown. 

%In addition to, the statistic analysis in Table \ref{tab: statistic about the extratced CS triplets in Commonsense-Dialogues dataset} also demonstrates that the extracted knowledge rarely includes the commonsense hidden in the response, which causes the response model prefer capturing the commonsense reasoning hidden in dialogue context rather than extensional knowledge that contains some keywords can further guide response generation.

The use cases in Table \ref{tab: use cases for the dialogpt+cs adapter generation mode} shows that even though the DialoGPT+CS\_Adapter response model generates only response without commonsense knowledge in some cases (the first example in Table \ref{tab: use cases for the dialogpt+cs adapter generation mode}), it is able to generate commonsense knowledge hidden in dialogue context along with a reasonable response in most cases (the second and third example in Table \ref{tab: use cases for the dialogpt+cs adapter generation mode}). Beyond that, the generated commonsense knowledge includes some key concepts (highlighted with \textcolor{red}{red} in Table \ref{tab: use cases for the dialogpt+cs adapter generation mode}) in some cases and guide the response generation to some extent.

\section{Conclusion and Future Work}
\label{sec: conclusion and future work}

In this work, we infuse commonsense knowledge into the pre-trained conversational model to enhance the commonsense capability. Hence, the commonsense guided dialogue response model can not only generate a response, but also the underlying commonsense triplets hidden in the dialogue interaction. To be more specific, we firstly integrate the commonsense knowledge in ConceptNet into pre-trained DialoGPT through CS\_Adapter fusion. Secondly, we utilize CommonGen for bridge dataset and propose ``two-way learning'' to train DialoGPT+CS\_Adapter for capturing the bidirectional relationship between commonsense triplets and sentence pairs. The Commonsense-Dialogues dataset is finally leveraged to further enable the commonsense knowledge understanding and reasoning for response generation.

The experiment results in Section \ref{subsec: Evaluation of Commonsense Adapter} demonstrate that the pre-trained conversation model DialoGPT benefits from the commonsense knowledge integration and possesses commonsense capability to generate series of reasonable commonsense triplets. The experiment results in Section \ref{subsec: Evaluation of Commonsense guided Response Model} show that our proposed DialoGPT+CS\_Adapter generation model can both generate commonsense reasoning along with a response. Based on the two-way learning with CommonGen dataset, the response model can generate better commonsense triplets that fits well to dialogue context. However, the generated responses have a little loss even compared with the DialoGPT baseline. The possible reason is the rule-based CS extraction method, which includes keywords extraction and knowledge extraction, does not consider the discourse information. Hence the extracted CS knowledge is kind of imperfect and does not better guide the response generation. In our future work, we will think about more how to extract more relevant commonsense knowledge hidden in dialogue interaction. Furthermore, we believe that the CS knowledge can guide an more informative response generation(like the response in Figure \ref{fig: The commonsense reasoning between the human-machine interaction.} includes more information compared with generic response ``cool.''). This could be verified in the future research. Another shortcoming we found in this work is that the relation distribution in ConceptNet is severely imbalanced which results in an over-generation of the “[related to]” relation.

%extracted commonsense triplets are not so perfect that can not better guide the response generation. In the future, we will think more about how to accurately extract the commonsense knowledge hidden in the dialogue interaction and better guide the model to generate the commonsense triplets that have more guidance over the response, rather than only the underlying commonsense knowledge hidden in dialogue history.

%\section*{Acknowledgements}

\bibliography{anthology,custom}

%\appendix
\section{Appendix}
\label{sec:appendix}
Several Tables are attached in the Appendix given the maximal $8$ pages limitation.
%\section{Appendices}
%The Table \ref{tab: mapping between assertion relation and natural language phrase} is the mapping between the \emph{relation} in ConceptNet and natural language (NL) phrase along with special character $[ ]$. Our experiment show that the special character $[ ]$ highly helps the model to distinguish the special relation and normal words.

\begin{table}
\footnotesize
\begin{center}
\begin{tabular}{cc}
\toprule
    relations in ConceptNet & NL phrases  \\
     
   \midrule
   RelatedTo  & [related to]    \\
   FormOf  & [form of]  \\
   IsA & [is a]   \\
   PartOf & [part of] \\
   HasA & [has a] \\
   UsedFor & [used for] \\
   CapableOf & [capable of] \\ 
   AtLocation & [typically located at] \\
   Causes & [causes] \\
   HasSubevent & [has subevent of] \\
   HasFirstSubevent & [begins with] \\
   HasLastSubevent & [concludes with] \\
   HasPrerequisite & [has prerequisite] \\
   HasProperty & [has property] \\
   MotivatedByGoal & [motivated by goal] \\
   ObstructedBy & [obstructed by] \\
   Desires & [desires] \\
   CreatedBy & [created by] \\
   Synonyms & [synonym] \\
   Antonyms & [antonym] \\
   DistinctFrom & [distinct from] \\
   DerivedFrom & [derived from] \\ 
   SymbolOf & [symbolically represents] \\
   DefinedAs & [defined as] \\ 
   MannerOf & [manner of] \\
   LocatedNear & [located near] \\
   HasContext & [used in context of] \\
   SimilarTo & [similar to] \\
   CausesDesire & [makes someone want] \\
   MadeOf & [made of] \\ 
   ReceivesAction & [receives the action of] \\
   
\bottomrule
\end{tabular}
\end{center}

\caption{\label{tab: mapping between assertion relation and natural language phrase} The mapping between assertion relations in ConceptNet and natural language (NL) phrases. The experiment shows that the special character $[ ]$ in NL phrases helps the model a lot to distinguish the relations from normal words.}
\end{table}

\begin{table*}
\footnotesize
\begin{center}
\begin{tabular}{cc}
    \toprule
    prompt input & <|commonsense|>: emailing [related to] email, email [related to] mail 
    \\ 
    %\hline
    \makecell{CS generation} & \makecell{mail [used in context of] computing, computing [related to] calculation, calculation [related to] calcul \\
    mail [related to] internet, internet [has property] useful, useful [related to] practical}  \\
    
    \midrule
    prompt input & <|commonsense|>: supercomputer [derived from] computer, computer [has property] expensive \\ 
    %\hline
    \makecell{CS generation} & \makecell{expensive [related to] price, price [related to] purchase, purchase [related to] acquire \\
    expensive [distinct from] inexpensive, inexpensive [related to] cheap}\\
    
    \bottomrule
\end{tabular}
\end{center}
\caption{\label{tab: use cases for the cs adapter integration} Use cases for Adapter integration show that the DialoGPT+CS\_Adapter is capable of generating series of commonsense triplets.}
\end{table*}

\begin{table*}
\footnotesize
\begin{center}
\begin{tabular}{cc}
    \toprule
    CS input & <|commonsense|>: audience [related to] speech; give [manner of] make, speech [related to] make ;
    \\ 
    %\hline
    \makecell{generated sentence} & \makecell{Celebrity giving a speech in the audience. \\
    A group of people give an audience a speech during an event.}  \\
    
    \hline
    CS input & <|commonsense|>: boat [related to] vessel, vessel [used for] float; boat [typically located at] river; 
    \\ 
    %\hline
    \makecell{generated sentence} & \makecell{A white boat floats in the water near the river. \\
    A boat floats in the river on a sunny day.}  \\
    
    \midrule
    \midrule
    sentence input & The dog eats food inside the house.\\ 
    %\hline
    \makecell{generated CS triplets} & \makecell{<|commonsense|>: dog [capable of] eat; eat [related to] food; \\
    <|commonsense|>: eat [related to] food; food [typically located at] house;}\\
    
    \hline
    sentence input & I laid the blanket on the bed. 
    \\ 
    %\hline
    \makecell{generated CS triplets} & \makecell{<|commonsense|>: bed [used for] sleep; blanket [typically located at] bed; bed [used for] lay; \\
    <|commonsense|>: bed [used for] lay; bed [related to] sleep, lay [related to] sleep;}  \\
    
    \bottomrule
\end{tabular}
\end{center}
\caption{\label{tab: use cases for the commongen bridge model} Use cases for DialoGPT+CS\_Adapter trained with extracted CommonGen in two-way method show the model can both generate sentences given commonsense triplets and underlying commonsense knowledge given a sentence.}
\end{table*}

\begin{table*}
\footnotesize
\begin{center}
\begin{tabular}{cc}
    \toprule
    dialogue context & Sorry I am late. \\ 
    CS reasoning $\&$ response & No worries. It's ok. We are here for you.\\
    
    \midrule
    dialogue context & I'm so tired. How long do i have to keep taking over your shift? \\ 
    CS reasoning $\&$ response & shift [is a] turn $\&$ It's only like 20 minutes, I think we can do this.\\
    
    \midrule
    dialogue context & I want to move back in with Casey. \\ 
    CS reasoning $\&$ response & \makecell[c]{move [is a] decision $\&$ I heard you were looking to move in with Casey.\\
    move [related to] \textcolor{red}{plan} $\&$ That would be a great \textcolor{red}{plan}.} \\

    \midrule
    dialogue context & hello friend, I got a mail that we have been hired for the movie. \\ 
    CS reasoning $\&$ response & movie [is a] \textcolor{red}{film} $\&$ I am so happy for you. I know you love to \textcolor{red}{film}.\\
    
    \midrule
    dialogue context & I have no idea how i did on that interview, yesterday, I was so nervous. \\ 
    CS reasoning $\&$ response & interview [related to] \textcolor{red}{job} $\&$ Why, did you not get the \textcolor{red}{job}? \\

   \bottomrule
\end{tabular}
\end{center}
\caption{\label{tab: use cases for the dialogpt+cs adapter generation mode} Use cases for final commonsense guided response generation model: DialoGPT+CS\_Adapter. The cases with highlighted \textcolor{red}{red} words that both exist in commonsense triplets and response show that the generated commonsense knowledge exerts guidance over the generated response and provides key concepts that occur in the generated response.}
\end{table*}

\end{document}